\documentclass[letterpaper, 10 pt, conference, twocolumn]{ieeeconf}

\IEEEoverridecommandlockouts

\usepackage{cite}
\usepackage{amsmath,amssymb,amsfonts}
\usepackage{multirow}
\usepackage{algorithm}
\usepackage{algorithmic}
\usepackage{graphicx}
\usepackage{textcomp}

\usepackage{multirow}
\usepackage[table,xcdraw]{xcolor}

\usepackage{soul}
\usepackage{subfig}

\usepackage{gensymb}

\def\BibTeX{{\rm B\kern-.05em{\sc i\kern-.025em b}\kern-.08em T\kern-.1667em\lower.7ex\hbox{E}\kern-.125emX}} 

\definecolor{deep-red}{RGB}{192, 0, 0}
\definecolor{deep-purple}{RGB}{120, 0, 170}
\definecolor{good-green}{RGB}{0,175,0} 
\definecolor{purple}{RGB}{210, 0, 210}

\makeatletter
\let\NAT@parse\undefined
\makeatother
\usepackage[colorlinks]{hyperref}

\begin{document}
\setlength{\textfloatsep}{8pt plus 2pt minus 4pt}

\title{FIRE-3DV: Framework-Independent Rendering Engine for 3D Graphics using Vulkan} 
\author{%
     Christopher John Allison$^{1}$, Haoying Zhou$^{3}$, Adnan Munawar$^2$, Peter Kazanzides$^2$, and Juan Antonio Barragan$^{2}$%
     \thanks{$^1$Khoury College of Computer Sciences, Northeastern University, Boston, MA 02115, USA.  
     Email: {\tt allison.ch@northeastern.edu}}
     \thanks{$^2$ Department of Computer Science, Johns Hopkins University, Baltimore, MD 21218, USA.  
     Email: {\tt jbarrag3@jhu.edu}}
     \thanks{$^3$ Department of Robotics Engineering, Worcester Polytechnic Institute, Worcester, MA 01608, USA.}
}

\maketitle
\begin{abstract}
Interactive dynamic simulators are an accelerator for developing novel robotic control algorithms and complex systems involving humans and robots. In user training and synthetic data generation applications, high-fidelity visualizations from the simulation are essential. Yet, robotic simulators often limit their rendering algorithms to preserve real-time interaction with the simulation. Advancements in Graphics Processing Units (GPU) enable improved visualization without compromising performance. However, these advancements cannot be fully leveraged in simulation frameworks that use legacy graphics application programming interfaces (API) to interface with the GPU. This paper presents a performance-focused and lightweight rendering engine supporting the modern Vulkan graphics API that can be easily integrated with other simulation frameworks to enhance visualizations. To illustrate the proposed method, our engine is used to modernize the legacy rendering pipeline of the Asynchronous Multi-Body Framework (AMBF), a dynamic simulation framework used extensively for interactive robotics simulation development. This new rendering engine implements graphical features such as physically based rendering (PBR), anti-aliasing, and ray-traced shadows, significantly improving the image fidelity of AMBF. Computational experiments show that the engine can render a simulated scene with over seven million triangles while maintaining GPU computation times within two milliseconds.
\end{abstract}

\section{Introduction}

Dynamic robotic simulators facilitate the development and testing of novel robotic control algorithms and complex systems involving humans and robots. In particular, interactive dynamic simulators have been used extensively as cost-effective and safe solutions to train human operators and machine learning algorithms in a wide range of domains \cite{thornblade_simulation-based_2021}. 
Simulation frameworks targeted for training applications require realistic visualizations of the workspace while maintaining real-time performance. In previous work, the Asynchronous Multi-Body Framework (AMBF) \cite{munawar_2019_ambf} was introduced as a general real-time dynamic simulator for robots and free bodies. This framework was developed by integrating several external tools, such as an extended version of CHAI-3D, BULLET-Physics, and OpenGL. With its flexible description format, AMBF has enabled the development of complex interactive simulations, such as an environment for learning robot-assisted surgical suturing \cite{munawar_2022_SRC} and a volumetric simulation for skull base surgery \cite{munawar_2024_volumetric_paper2}. 

Currently, AMBF rendering and computing capabilities are limited by the legacy OpenGL graphics application programming interface (API) at the core of its rendering pipeline. AMBF relies on OpenGL version 2.1, released in 2006, to perform graphics operations. This severely limits AMBF's access to new developments in GPU hardware and real-time rendering that improve performance and visual fidelity \cite{fried2004proving ,agha2015role}. A particular limiting factor of legacy OpenGL is the lack of support for general-purpose computing on the GPU, a groundbreaking development that can greatly accelerate computationally intensive processes, such as soft-body physics simulation, ray-tracing, and deep learning \cite{bailey2016computeopengl}. Other simulation engines dependent on OpenGL face similar issues and will eventually need to pivot to a modern graphics API to stay compatible and performant on modern hardware.

\begin{figure}[ht]
    \centering
    \includegraphics[width=0.48\textwidth]{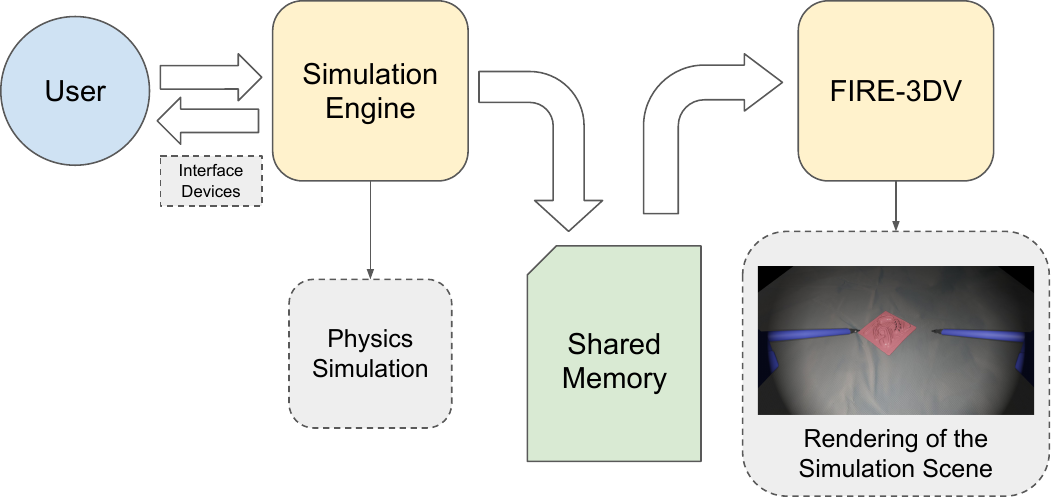}
    \caption{Overview of the proposed dynamic simulation framework. The simulation engine handles user input through interface devices and runs physics simulations on the scene, recording scene updates in shared memory. FIRE-3DV reads the scene updates from shared memory and renders the result with high visual fidelity.}
    \label{fig:fig system overview}
\end{figure}

This work presents a framework-independent rendering engine for 3D graphics using Vulkan (FIRE-3DV) to modernize AMBF and other simulators limited by legacy rendering engines. We chose Vulkan, OpenGL’s successor, due to its cross-platform support and low-level design, allowing for significant performance improvements.  Given the substantial architectural differences between legacy OpenGL and Vulkan, it was decided to develop the renderer as a standalone program and, therefore, integrate both processes via a plugin in the simulation engine. As such, support for interface devices, physics computations, and plugin architecture is maintained in the simulation engine, while FIRE-3DV only takes care of the scene visualization (see Fig. \ref{fig:fig system overview}). Overall, the full system retains access to the simulation engine's features while significantly improving rendering quality.

While developing this rendering engine, efforts were concentrated on improving the computational performance of the rendering pipeline compared to AMBF. Additionally, several additional features, such as support for physically-based rendering (PBR) materials, anti-aliasing, and ray-traced shadows, were co-developed to improve the visual fidelity of the engine. Overall, this paper summarizes some of the main design principles and challenges faced while developing this system, as well as opportunities for using a modern graphics API in developing advanced dynamic simulations. In summary, our key contributions are:

\begin{enumerate}
    \item FIRE-3DV, a standalone and open-source rendering engine supporting the Vulkan graphics API. 
    \item Integration between the AMBF dynamic simulation engine and FIRE-3DV, as an illustration of how this approach could be used with other simulators.
\end{enumerate}



\section{Related Work}
Several commercial and open-source dynamic robotic simulators use a modular software design that enables integration with different rendering engines. For instance, Gazebo\cite{koenig2004design} uses the OGRE rendering engine by default but also provides experimental support for Nvidia Optix. Similarly, the SOFA simulator \cite{faure2012sofa} supports several rendering options such as Unity, Godot, and Unreal engine. In this work, it was decided to build a customized rendering engine rather than using available all-purpose rendering engines to upgrade the AMBF simulation. Some advantages of creating a new engine instead of using existing ones are optimized rendering performance as the render only needs to support material models relevant to AMBF simulations, application-specific optimizations such as disabling frustum culling in low geometry scenes, and simplified development of application-specific features. 

For AMBF, our previous work \cite{barragan2024_realistic} presented a plugin that enabled rendering AMBF scenes using Blender's built-in rendering engines, EEVEE and Cycles\cite{astuti2022comparison}, and used the Robotic Operating System (ROS) \cite{quigley2009ros} as a communication layer. This plugin permitted AMBF to use all rendering features provided by Blender; however, it could not render scenes at interactive rates due to communication latency and the processing overhead of Blender's interface. Compared to our previous work, the proposed rendering engine presents several optimizations that simultaneously improve visual fidelity and rendering speed. The systems's overall performance can be attributed to the use of a modern graphics API and a lightweight communication layer between AMBF and the proposed rendering engine. 

The Vulkan API caught our attention due to its great performance on ray-tracing \cite{souza2021analysis, saed2022vulkan}, and complex shading advancements \cite{danliden2020multi} that could facilitate advanced visual effects such as light reflection and detailed material textures.  Furthermore, the Vulkan API is cross-platform \cite{sellers2016vulkan, ioannidis2020multithreaded} and facilitates the use of ray-tracing acceleration units\cite{montoto2021hands}. Vulkan has been increasingly implemented in recent academic research \cite{unterguggenberger2023vulkan}. Taking advantage of the low-level API and integration compatibility, Vulkan has been deployed into various domains: physics particle simulation \cite{zverev2021development, ozvoldik2021assembly}, efficient data visualization \cite{rossant2021high}, deformable body simulation \cite{kamnert2023real}, and surgical simulations \cite{xu2021surrol, maul2021x}. Given all the advantages presented, Vulkan is an excellent option for modern rendering engines. 

\begin{figure*}[ht]
    \centering
    \includegraphics[width=1.0\textwidth]{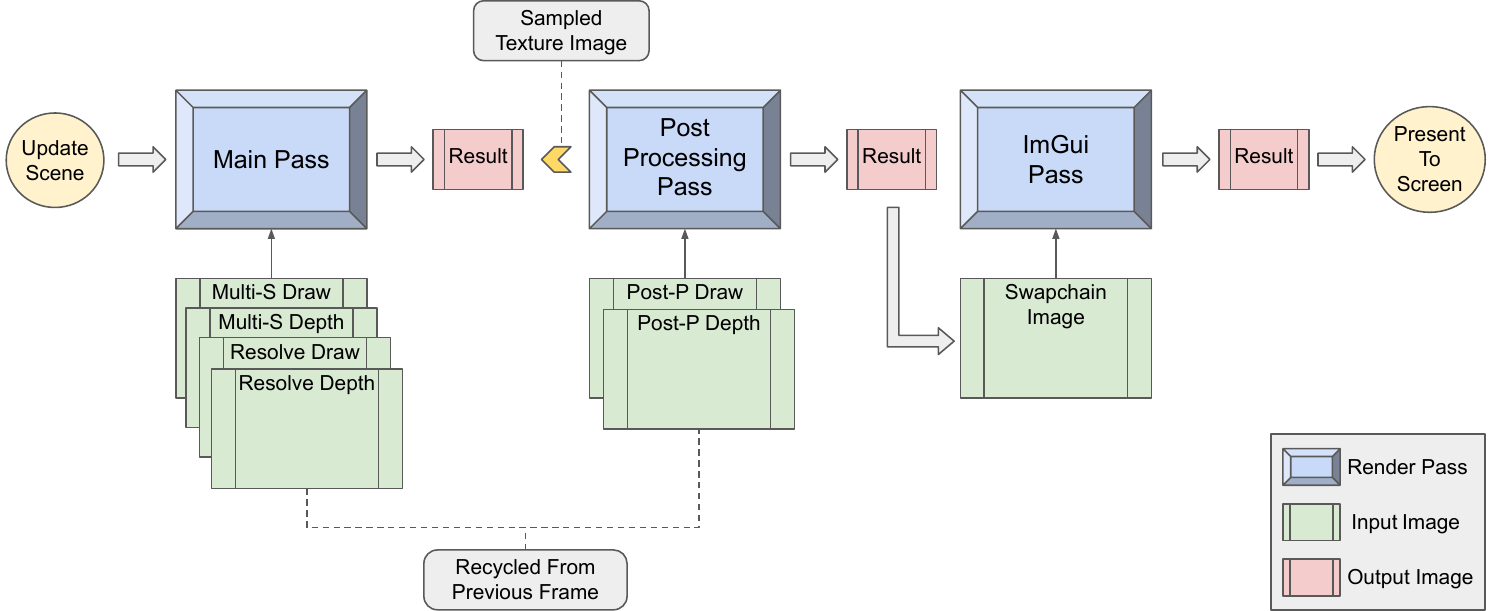}
    \caption{Overview of FIRE-3DV's render loop from the scene update to the presentation of the image to the screen. The scene geometry is rendered in the Main Pass, and its resulting image is used as a sampled texture by the Post-Processing Pass. The ImGui pass renders the GUI overtop the image and the result is presented to the screen.}
    \label{fig:RenderLoop}
\end{figure*}
\section{Background}
\subsection{Overview of graphics APIs and OpenGL}
A graphics API provides a set of standardized commands and functions that allow developers to communicate with a computer’s GPU.  OpenGL is a graphics API created by the Khronos Group that has had a rich history of development since 1992. Pertinent to this paper is the fact that OpenGL versions lower than 3.0 are referred to as legacy OpenGL \cite{wright2007opengl}, as they use a deprecated fixed function pipeline. Legacy OpenGL was designed before the advent of programmable shaders and is characterized by computationally expensive CPU operations, e.g., immediate mode vertex attribute specification. In addition to reduced performance due to CPU overhead, legacy OpenGL has limited support for modern graphics debuggers and profilers, such as RenderDoc \cite{castorina2023mastering} and NVIDIA Nsight Graphics \cite{iyer2016gpu}.


Modern OpenGL (versions 3.0 to 4.6) progressively shifted towards a more programmable graphics pipeline, adding support for features like programmable shaders, compute shaders, and multi-threaded command buffer submission, among others. In 2017, active development of OpenGL was ceased in favor of the newer Vulkan API, making OpenGL 4.6 the final version. 
This is relevant as dynamic simulators relying on OpenGL will never receive support for new GPU features, such as hardware-accelerated ray-tracing. Moreover, GPU manufacturers and driver developers may decide to stop supporting the OpenGL specification, as seen by Apple deprecating OpenGL on all of their platforms in 2018. Overall, it was decided to support Vulkan instead of a newer OpenGL version to ensure long-term support for the rendering engine on modern hardware.
\subsection{Overview of the Vulkan Graphics API}
Modern graphics APIs, such as Vulkan, DirectX12, and Metal, choose to put more control in the hands of the user rather than the driver. This leads to performance improvements at the cost of a more verbose and complex API. Vulkan requires explicit management of GPU memory resources, synchronization objects, debug messaging, context initialization, and more. An additional benefit of this low level of abstraction over the GPU is that modern graphics APIs are more similar, and therefore, rendering engines can more easily support additional APIs. Overall, Vulkan and other modern graphics APIs allow for programming performant programs tailored to the application's specific needs.
\subsection{Overview of AMBF}  AMBF integrates several external tools to enable interactive dynamic simulations. Rendering is enabled in AMBF by an extended version of CHAI-3D \cite{conti2003chai} (actively developed along with AMBF) that uses the legacy OpenGL version 2.1 at its core. CHAI-3D is also utilized to interface with haptic devices. Dynamic computations are performed by BULLET-Physics  \cite{coumans2015bullet}. Assets in AMBF are specified using AMBF description files (ADF), which define the properties of simulated bodies in a human-readable format. Lastly, AMBF facilitates the creation of ADF files by using the AMBF Blender-addon \cite{ambfblender_addon}, a plugin that permits Blender to export objects into ADFs. 
\section{Methodology}
The complete dynamic simulation framework is composed of three different components (see Fig. \ref{fig:fig system overview}). Firstly, we have the standalone rendering engine Framework-Independent Rendering Engine for 3D graphics using Vulkan (FIRE-3DV). By itself, FIRE-3DV only renders the loaded objects and provides the option to move the virtual cameras and adjust visual parameters via a GUI. Secondly, we have the simulation engine running in headless mode to perform physics computations and communicate with user input devices. Thirdly, we have a memory region shared between both processes that is used to record scene updates given by the simulation engine to update the scene in FIRE-3DV. Overall, these three components work together to enable the user to access the full set of features from the dynamic simulator, while also obtaining high-quality renderings from FIRE-3DV. The rest of the methodology is divided as follows. Subsection \ref{subsect: AMBF_vulkan engine} describes the software design of the renderer FIRE-3DV. Subsection \ref{subsect: ambf-vulkan graphical} presents the graphical features that were implemented in FIRE-3DV to improve visual fidelity. Lastly, subsection \ref{subsect: interchange with AMBF} describes the interprocess communication plugin for AMBF and FIRE-3DV as example for how other simulators could integrate with FIRE-3DV. 
\subsection{FIRE-3DV rendering engine architecture} \label{subsect: AMBF_vulkan engine}
\begin{figure*}[ht]
    \centering
    \subfloat[Aluminum]
    {
    \includegraphics[height=0.15\linewidth]{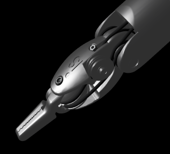}
    \label{fig:PBR-a}
    }
    \hfill
    \subfloat[Gold]
    {
    \includegraphics[height=0.15\linewidth]{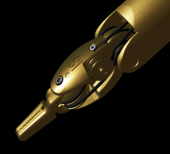}
    \label{fig:PBR-b}
    }
    \hfill
    \subfloat[Copper]
    {
    \includegraphics[height=0.15\linewidth]{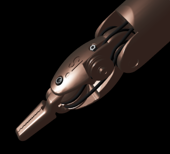}
    \label{fig:PBR-c}
    }
    \hfill
    \subfloat[Glossy Plastic]
    {
    \includegraphics[height=0.15\linewidth]{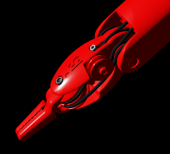}
    \label{fig:PBR-d}
    }
    \hfill
    \subfloat[Matte Plastic]
    {
    \includegraphics[height=0.15\linewidth]{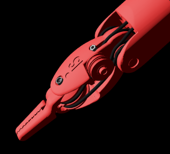}
    \label{fig:PBR-e}
    }
    \caption{Material examples using the PBR reflection model. For (a)-(c), normal incidence Fresnel reflectance measurements are used to create physically based metallic materials. For (d)-(e), the roughness parameter allows for distinction between glossy and matte materials.}
    \label{fig:PBR}
\end{figure*}
\subsubsection{Render Loop Overview}
Figure \ref{fig:RenderLoop} illustrates a high-level overview of the rendering loop implemented in FIRE-3DV. The data flow depicted is as follows: (1) The scene is updated to reflect the changes made from device input and model transformation updates (see \ref{subsect: interchange with AMBF}), (2) mesh draw commands are sorted and submitted for rendering, (3) post-processing operations are applied to the main pass' output, (4) the Dear ImGui \cite{knoblauch2017imgui} library renders an immediate-mode Graphical User Interface (GUI) onto the post-processed image, and (5) the final frame is presented to the screen.

Key to the performance of our rendering engine is its implementation for storing scene assets and submitting GPU draw commands. FIRE-3DV opts to store model resource handles in global type-specific containers. The scene data is organized by a nodal hierarchy, with each node pointing to the resources it uses. This way, we can create a collection of draw command requests containing the necessary resource information and sort the list to optimize draw performance. We accomplish this by sorting by material type and vertex buffer index, which minimizes the amount of descriptor (a modern graphics API term to describe bound shader resources, e.g., textures or uniform buffers) bind commands and vertex buffer cache misses.

With the draw commands sorted, we can submit the Main Pass command buffer to the GPU. The vertex shader of the main pass is responsible for transforming the vertices into view space and passing shading data to the fragment shader. FIRE-3DV uses a performance-improving technique called vertex pulling: rather than additionally binding vertex buffers for each draw, we store the vertex information in a buffer and index into it, like an array. The fragment shader is responsible for lighting calculations discussed in detail in \ref{subsect: ambf-vulkan graphical}. Additionally, the fragment shader converts color data from sRGB gamma space to linear space to perform lighting calculations. Afterward, we apply Reinhard tone mapping to map color values greater than 1.0 to a value between 0.0 and 1.0. Otherwise, they will be clamped to 1.0 automatically. Finally, we invert the aforementioned linear space conversion to bring the colors back into sRGB gamma space, an operation known as gamma correction. 

The image result of the Main Pass is then converted for use as a texture image in the following pass, the Post-Processing Pass. This render pass is typically used to perform image processing techniques on a rendered image. In FIRE-3DV, we currently use the Post-Processing Pass to perform a screen-space anti-aliasing algorithm discussed in \ref{subsect: antialiasing}. The vertex shader creates a screen-sized quad and passes its UV indices to the fragment shader. The fragment shader samples the Main Pass resulting texture and performs the anti-aliasing algorithm. 

Lastly, the result of the Post-Processing Pass is converted into an image format matching that of the display device. The blitted image is then used to render a GUI over the scene using the Dear ImGui library. Rather than a retained-mode GUI, which can add a lot of unnecessary complexity due to state management, we opted for an immediate-mode GUI. While this choice adds unnecessary drawing of unchanged GUI state, the cost is negligible for our purposes, as seen in Figure \ref{fig:VulkanNsight}. In return, we get the flexibility and simplicity of the Dear ImGui library for our debugging and scene customization needs. Currently, the GUI displays a window listing basic debugging and profiling information, e.g., frame time and camera coordinates. With the ImGui Pass finished, the final image is ready to be presented to the screen.
\subsubsection{System Improvements}
Rather than an uncapped frame rate, the system synchronizes with the display device, limiting the frame rate to the refresh rate of the display device. This prevents unnecessary computation while allowing the rate cap to be display-dependent, which can be beneficial if the simulation is displayed via a virtual reality headset. FIRE-3DV uses a double-buffered swapchain, i.e., while one frame is rendered by the GPU, the other is prepared for rendering by the CPU. As shown in Figure \ref{fig:RenderLoop}, render attachments, the resources that the render pass writes to, are recycled from the previous frame. In actuality, this means two frames prior due to the double buffer.

The system uses deletion queues to manage resource lifetimes. There are three deletion queues in total: the main deletion queue and one queue for each frame of the double-buffered swapchain. The main deletion queue is flushed during application shutdown, freeing objects that persist over the entire duration of the program. Frame-specific deletion queues are flushed after the corresponding swapchain image is finished presenting to the screen. These deletion queues trivialize resource lifetime management, reducing it down to pushing back a lambda function to the necessary queue.
\subsection{Graphical features of FIRE-3DV} \label{subsect: ambf-vulkan graphical}
A central motivation for building a modernized rendering engine for dynamic simulation was to surpass the graphical quality restrictions of our previous implementation, AMBF. In the context of surgical simulations, realistic visualizations play an important role in both operator training and machine learning. The following section describes the graphical improvements that were implemented in FIRE-3DV.
\subsubsection{Physically Based Rendering}
Physically Based Rendering (PBR) is a lighting model approach in Computer Graphics that enables photo-realism through approximation of real-world physics\cite{whitted2005_improved}. This is especially useful in a simulation environment where real-world measurements can be used to create a more accurate representation. Figures \ref{fig:PBR-a},  \ref{fig:PBR-b} and \ref{fig:PBR-c} demonstrate how we can use normal incidence Fresnel reflectance measurements to shade metallic materials more accurately. Figures \ref{fig:PBR-d} and \ref{fig:PBR-e} demonstrate different roughness values on a non-metallic material. FIRE-3DV uses a Cook-Torrance specular reflectance model variation that is common in real-time rendering applications \cite{cook1982reflectance}:
\[f_s = \frac{D(\mathbf{n}, \mathbf{h}, \alpha)F(\mathbf{h}, \boldsymbol{\omega}_i, F_0)G(\mathbf{n}, \boldsymbol{\omega}_o, \boldsymbol{\omega}_i, \alpha)}{4(\boldsymbol{\omega}_o \cdot \mathbf{n})(\boldsymbol{\omega}_i \cdot \mathbf{n})}\]

\noindent
where \(f_s\) is the specular bidirectional reflectance distribution function, the chosen normal distribution function \(D\) is Trowbridge-Reitz GGX\cite{walter2007microfacet}, the chosen Fresnel function \(F\) is the Schlick approximation\cite{schlick1994inexpensive}, the chosen geometry function \(G\) is Schlick-GGX, \cite{karis2013real}, \(\mathbf{n}\) is the normal vector, \(\mathbf{h}\) is the halfway vector, \(\boldsymbol{\omega}_o\) is the normalized incoming light direction, \(\boldsymbol{\omega}_i\) is the normalized outgoing light direction, \(F_0\) is the normal incidence Fresnel reflectance, and \(\alpha \) is the surface roughness value. This variation was chosen due to its computational efficiency in real-time applications and compatibility with glTF's metallic-roughness workflow.

Previous reflection models, such as Blinn-Phong\cite{blinn1977models}, create physically implausible shading due to their inability to conserve energy and represent microscopic surface changes. Additionally, its material parameters, such as ``Shininess,'' cannot be converted from physical measurements. PBR allows photogrammetric measurements of material properties, such as albedo and surface roughness, to accurately replicate a material's reflectance in a simulation environment.

\subsubsection{Anti-Aliasing} \label {subsect: antialiasing}
\begin{figure}[ht]
    \centering
    \includegraphics[width=.4\textwidth]{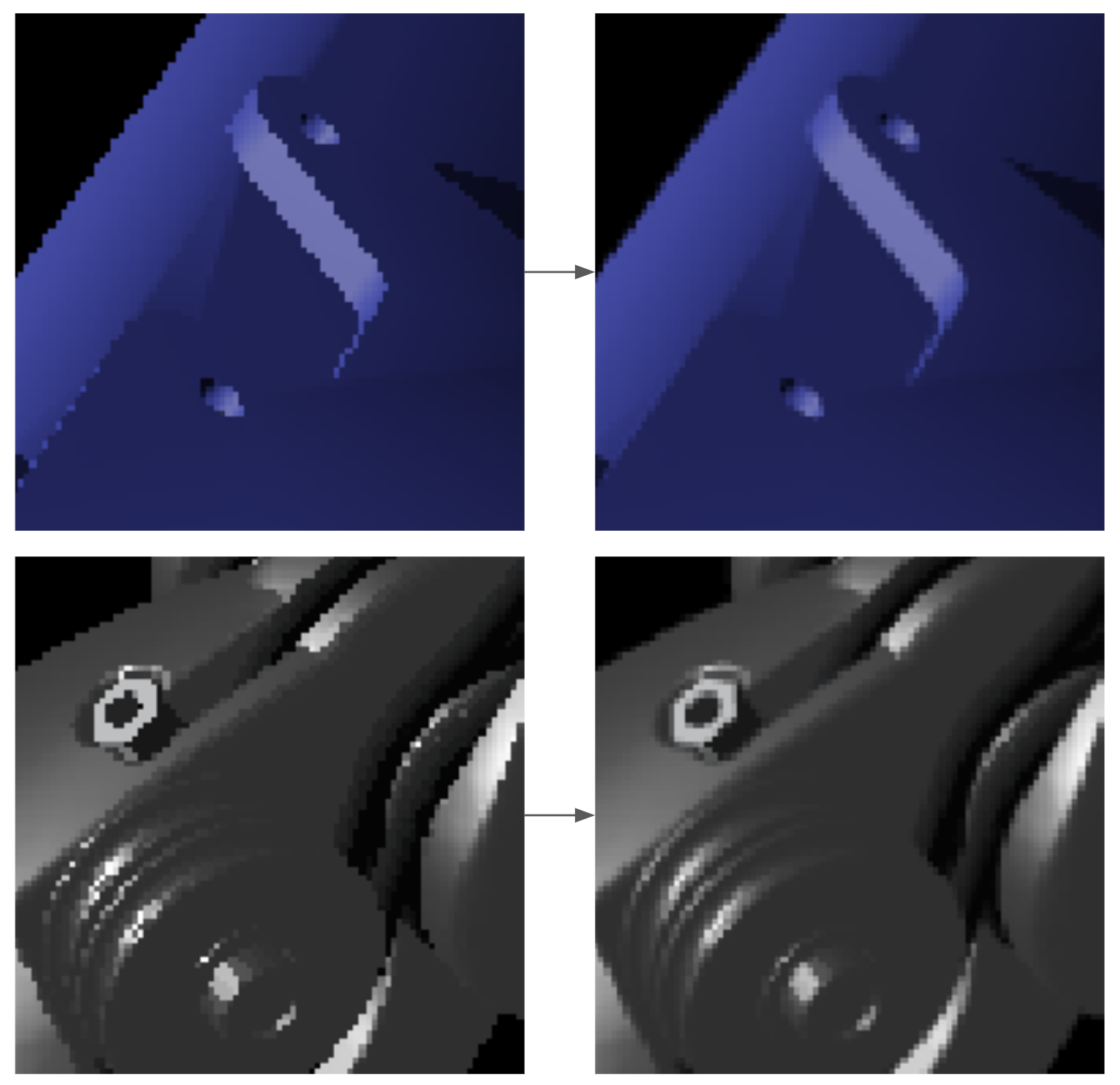}
    \caption{Comparison between anti-aliasing disabled (left) and enabled (right). Aliasing artifacts are due to undersampling of the scene in high-contrast areas. The top images show an improvement in edge jaggedness. The bottom images show an improvement in specular highlight clarity.}
    \label{fig:Anti-Aliasing}
\end{figure}

FIRE-3DV employs two techniques to reduce visual artifacts caused by under-sampling the scene (see Figure \ref{fig:Anti-Aliasing}). These techniques differ in the types of aliasing they target, as well as the stage of rendering they are implemented in, and we combine them to improve image quality further. 

Multisample Anti-Aliasing (MSAA) allows for additional sampling per pixel with little performance cost. Rather than shading each additional sample, as with Supersampling Anti-Aliasing, additional samples are only shaded if the triangle they intersect is not oversampled within the pixel. Before beginning the main render pass, FIRE-3DV submits a multisample draw and depth image, along with an additional resolve draw and depth image, and the GPU will automatically handle the rest.

Fast Approximate Anti-Aliasing (FXAA)\cite{jimenez2011filtering} is a post-processing technique that reduces aliasing without any extra sampling of the scene. Using high changes of luminance from nearby pixels, the algorithm detects edges of geometry and subsequently blends along the edge. Visual improvements are minor compared to MSAA but are still worth including due to the technique's low computational cost.

\subsubsection{Ray-traced shadows}
\begin{figure}[ht]
    \centering
    \includegraphics[width=0.4\textwidth]{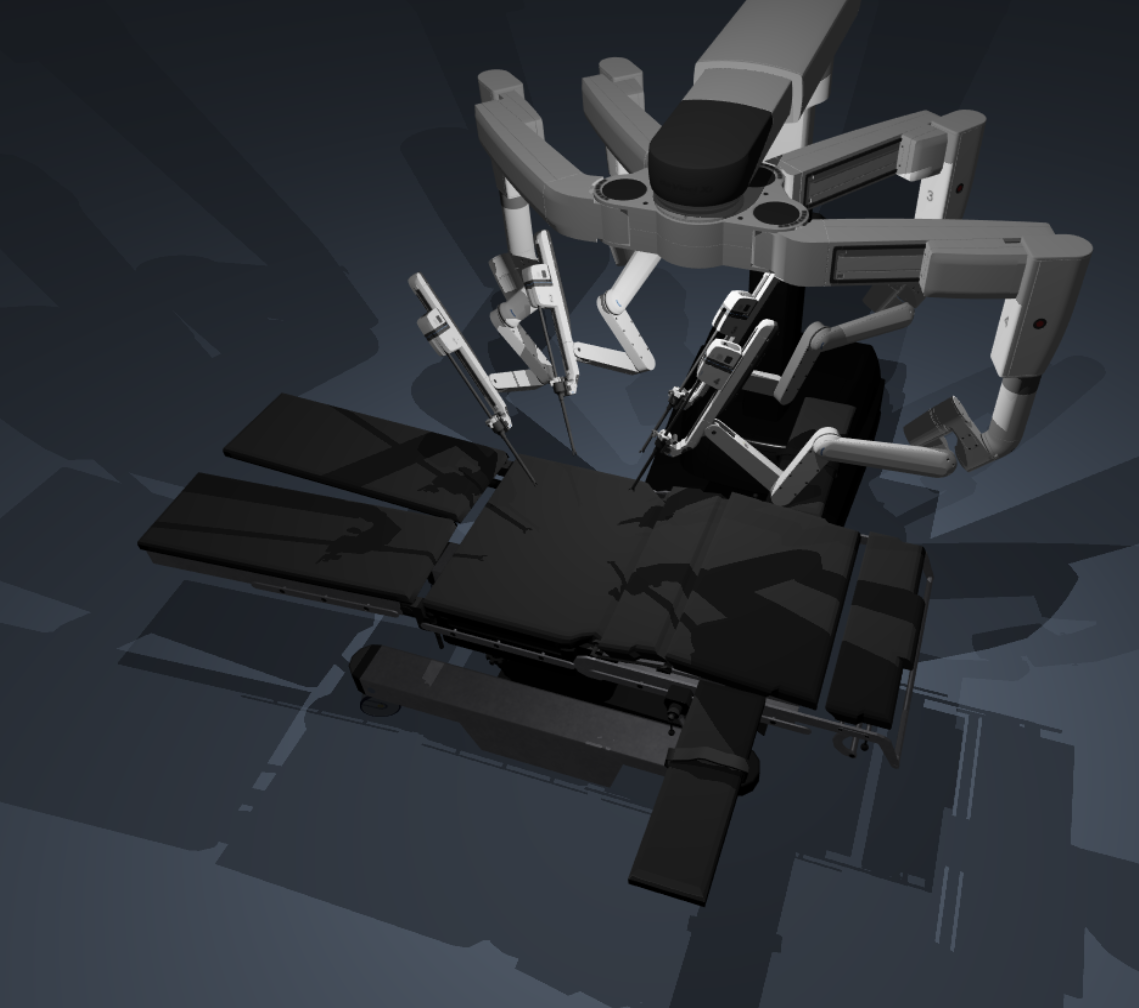}
    \caption{Demonstration of the engine's ray-traced omnidirectional shadows in a scene with two light sources. A ray query traverses the acceleration structure from the fragment shader to determine the light visibility of the sampled point in the scene.}
    \label{fig:Shadows}
\end{figure}

GPU ray-tracing support is a clear example of the benefits of moving to a modern graphics API such as Vulkan. Similar to PBR, ray-tracing approximates real-world light transport physics, resulting in more photorealistic images. Due to its high computational expense, ray-tracing in real-time applications was limited until recent advancements in hardware acceleration and denoising techniques. Using Vulkan's ray-tracing extensions, we implemented dynamic ray-traced hard shadows as an alternative to shadow mapping techniques (See Figure \ref{fig:Shadows}). 

Algorithms designed to utilize ray-tracing acceleration hardware require the implementation of an acceleration structure build system. In short, acceleration structures enable the computation of ray intersections with scene geometry in logarithmic time complexity and are essential for real-time ray-tracing. Once the acceleration structures are set in place, we can use the \texttt{GLSL\_EXT\_ray\_query} extension from the fragment shader to ray-trace the scene and get important information, such as whether or not a pixel is in shadow. To determine if a pixel is in shadow, a ray is traced from the fragment's world space coordinate to the light's world space coordinate. If the ray hits any opaque geometry during the tracing, then the fragment is in shadow and does not receive direct light.

To build the acceleration structures in our engine, the following steps are performed: (1) on scene initialization, the geometry of each unique mesh is built into a corresponding GPU-side buffer called a Bottom Level Acceleration Structure (BLAS), (2) each BLAS is subsequently compacted for faster traversal and a smaller memory footprint, (3) an Instance object is built for each mesh instance, recording its initial transformation matrix, (4) the transforms are updated on each frame's scene update, and a single Top Level Acceleration Structure (TLAS) is built, containing each Instance. Because the application does not yet support geometry deformation, updating or rebuilding the BLASs is unnecessary. Furthermore, because TLAS updates deteriorate scene traversal quality, we choose to rebuild the TLAS every frame for a net increase in performance. Additionally, we offload all acceleration structure build tasks to an async compute queue, virtually removing the performance cost.
\subsection{Plugin communication between AMBF and FIRE-3DV} \label{subsect: interchange with AMBF}
\begin{figure}[ht]
    \centering
    \includegraphics[width=0.5\textwidth]{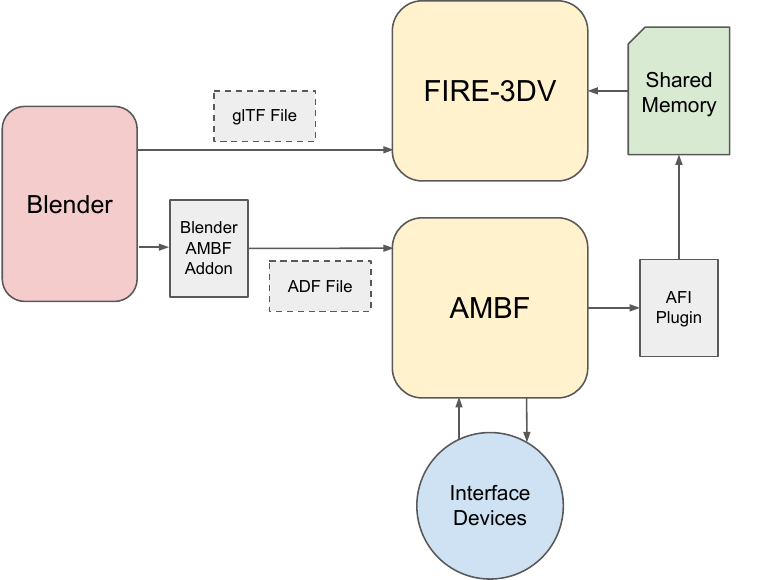}
    \caption{Overview of the AMBF-FIRE-3DV interchange system. The scene is exported from Blender and then imported into AMBF and FIRE-3DV. The AFI plugin records the transform matrices from all the objects inside AMBF and writes them into shared memory at every physics update. FIRE-3DV reads from shared memory to update each model's pose before rendering.}
    \label{fig:Interchange}
\end{figure}

To obtain a complete dynamics simulator, the rendering engine FIRE-3DV requires a physics engine that can update the poses of all the bodies in the simulation. In this work, we choose the physics engine inside AMBF, but the software design is flexible enough that other engines could be used. To enable communication between AMBF and FIRE-3DV, we designed the AMBF-FIRE-3DV Interchange (AFI) plugin, which enables AMBF to write the pose of all the bodies at every physics iteration into shared memory. FIRE-3DV then uses the poses in shared memory before rendering the scene (see Fig. \ref{fig:Interchange}).
AFI uses the \textit{boost::interprocess} library for a fast, lightweight, and operating-system-agnostic solution for interprocess communication. We simply use AFI to record AMBF's copy of each mesh node's world-space transformation matrix in a shared memory region, and FIRE-3DV updates its own copy of the mesh node's matrix. This one-way data transfer approach avoids many of the pitfalls of concurrency. Additionally, the library's \textit{managed\_shared\_memory} class will internally handle the ownership swapping of the shared memory region using a recursive mutex. Lastly, we chose to simply overwrite each matrix every frame and read each matrix every frame, regardless of whether or not the data has changed. We can afford to use this inefficient but error-resistant implementation because of the relatively low amount of meshes in a typical AMBF simulation environment.
\subsection{Integration with other dynamic simulation engines}
Similarly to AMBF, other 3D dynamic simulators can intercommunicate with FIRE-3DV to achieve higher-quality visual results with little modification to the simulation engine. Using a plugin architecture or inserting a small amount of code into the render loop, simulation engines must write each models transformation matrix into shared memory every frame, allowing FIRE-3DV to update the poses. Optionally, the simulation engine can write a change of basis matrix into shared memory to account for differences in bases between the simulation engine and FIRE-3DV. If the simulation engines uses a different file format than glTF, the scene can be exported to glTF using a tool like Blender, enabling FIRE-3DV to load the scene.
\section{Experimental results}
To evaluate the computational performance of the complete simulation framework, AMBF and FIRE-3DV were run on an Ubuntu 22.04 workstation with an NVIDIA GeForce RTX 4090 GPU. Two Blender scenes were used to evaluate the frameworks. Scene 1 includes two sets of da Vinci Patient Side Manipulators (PSMs) and a training suturing phantom taken from \cite{munawar_2022_SRC, barragan2024_realistic}, and has a total of 223K triangles. Scene 2 includes a model of a full da Vinci Si that was acquired from DistroSquid\footnote{\url{https://www.turbosquid.com}} and has a total of 550K triangles. Each scene was exported from a Blender file into ADF and glTF formats and then loaded to AMBF and FIRE-3DV, respectively. For every test, all the geometry was kept within the clipping plane, and no culling operations were applied to reduce the triangle count.
\subsection{Computational performance of FIRE-3DV}
\begin{figure}[ht]
    \centering
    \includegraphics[width=0.5\textwidth]{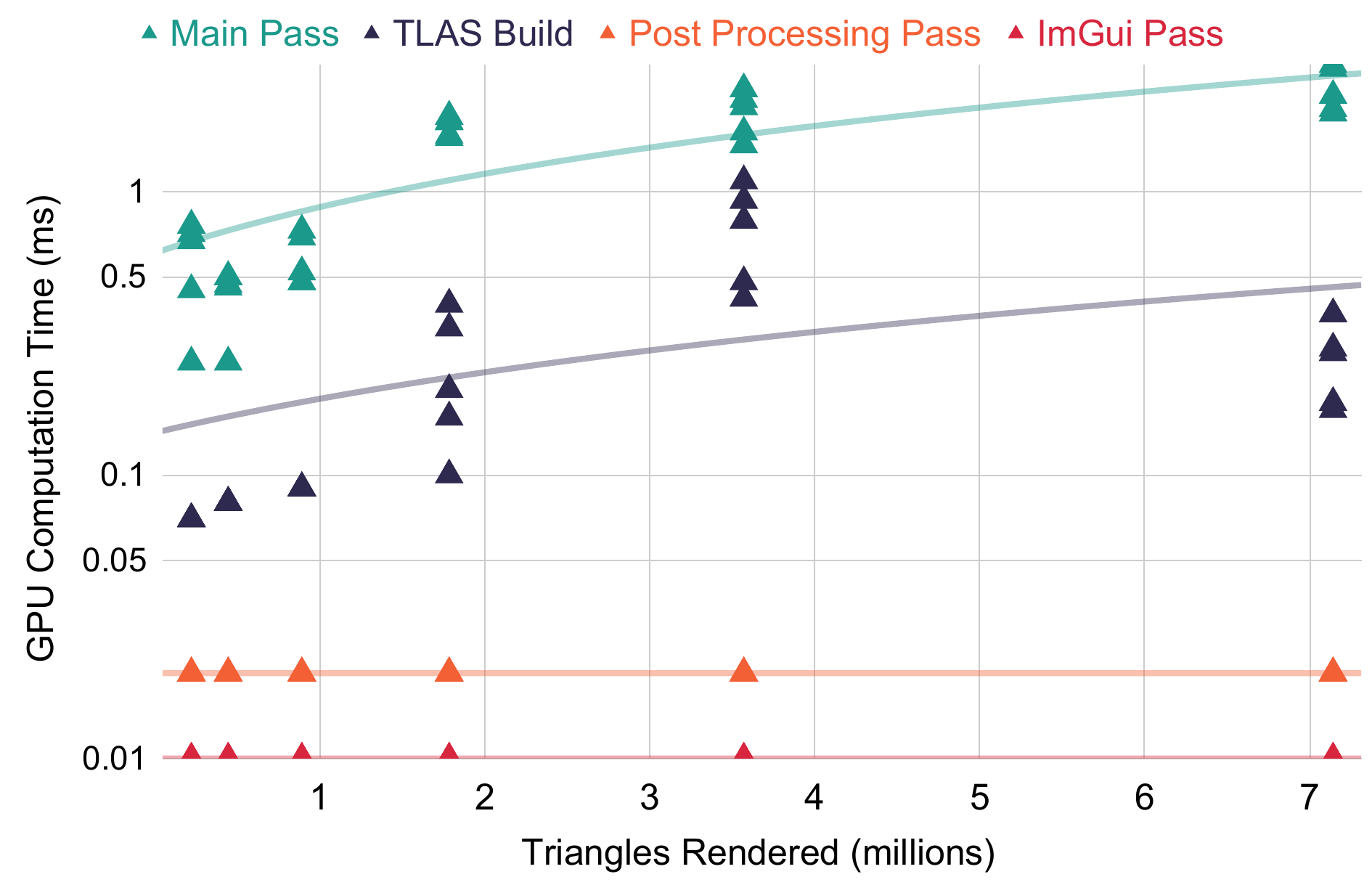}
    \caption{GPU computation time measurements of segments of FIRE-3DV's render pipeline. Measurements were taken from scenes comprised of increasingly greater triangle counts.}
    \label{fig:VulkanNsight}
\end{figure}

Figure \ref{fig:VulkanNsight} shows the GPU processing time for each stage in the rendering pipeline of FIRE-3DV as the total number of triangles increases. To easily augment the number of triangles, we duplicated the objects in Scene 1 so that the number of triangles would grow exponentially at every test. Using NVIDIA Nsight Graphics GPU Trace, five frame captures of each scene were profiled. The computational times for the following tasks were recorded: TLAS building, the main render pass, the post-processing pass, and the ImGui pass.  
It is observed that GPU computation time for building the TLAS and the main pass increases with the number of triangles. Conversely, the post-processing pass and ImGui pass remain constant throughout the experiment. An increase in TLAS building time is related to the increasing number of instances the TLAS represents, resulting from the rising model count. Similarly, an increase in the main pass is also reasonable as an increasing number of vertices directly relates to the number of  vertex shader invocations and triangles to rasterize. The post-processing and the ImGui pass manipulate the resulting image from the main pass and are, therefore, unaffected by an increase in geometry. Post-processing passes would only require more GPU computation time if the screen resolution increases or more complex image processing techniques are used. Overall, even with a scene with more than 7 million triangles, our proposed rendering engine maintains GPU computation times within 2\,ms. 
\subsection{Quantitative comparisons between AMBF and FIRE-3DV}
\begin{table}[]
\centering
\begin{tabular}{|c|cc|cc|}
\hline
Engine  & \multicolumn{2}{c|}{AMBF}             & \multicolumn{2}{c|}{FIRE-3DV}          \\ \hline
        & \multicolumn{1}{c|}{mean (ms)} & std  & \multicolumn{1}{c|}{mean (ms)}     & std  \\ \hline
Scene 1 & \multicolumn{1}{c|}{3.34}      & 0.68 & \multicolumn{1}{c|}{\textbf{0.55}} & 0.01 \\ \hline
Scene 2 & \multicolumn{1}{c|}{4.74}      & 0.16 & \multicolumn{1}{c|}{\textbf{0.56}} & 0.00 \\ \hline
\end{tabular}
\caption{Mean GPU usage on the render loop's main pass in
AMBF and FIRE-3DV measured in ms.}
\label{tab:table gpu usage}
\end{table}
Table \ref{tab:table gpu usage} shows the main pass GPU computation time of both FIRE-3DV and AMBF while rendering scene 1 and scene 2. Similarly to the previous test, NVIDIA Nsight Graphics GPU Trace was used to capture twenty frames of each scene for each engine. For each frame capture, we measured the computational time for the main render pass, which only consists of the drawing and shading of the scene's geometry.  

The results show that FIRE-3DV has an average speed increase of 6.1 and 8.5 times over AMBF for scene 1 and scene 2, respectively. Additionally, FIRE-3DV's measurements showed less variance. AMBF's computation times increased from scene 1 to scene 2, which is expected as scene 2 contains more triangles and more textures to render. On the other hand, FIRE-3DV used roughly the same computation time for both scenes. This is likely due to the GPU computation being so fast that the application is CPU-bound.
\subsection{Qualitative comparisons between AMBF and FIRE-3DV}
\begin{figure}[ht]
    \centering
    \includegraphics[width=0.5\textwidth]{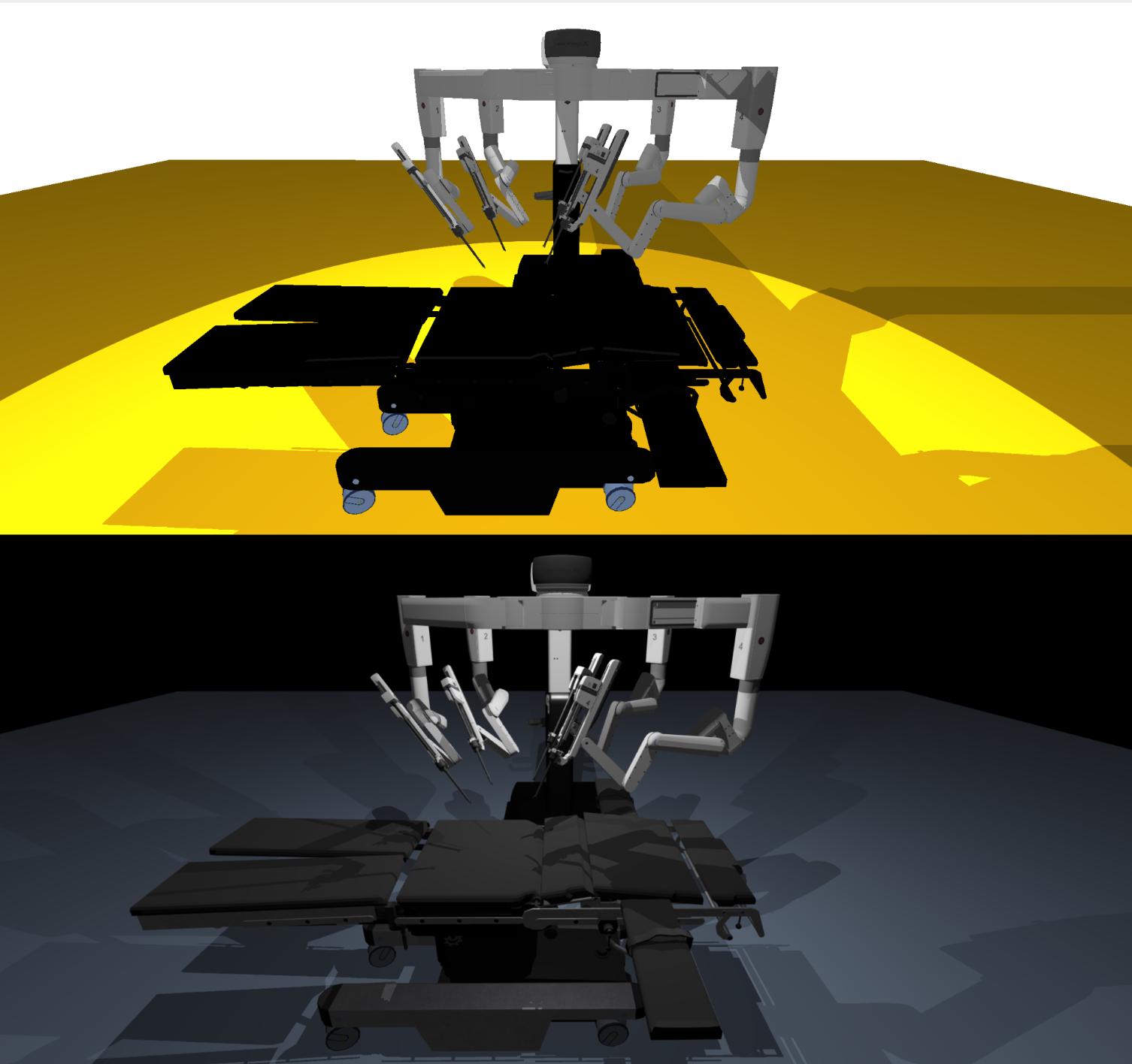}
    \caption{Scene 2 rendered from the same viewpoint in AMBF (top) and FIRE-3DV (bottom) for qualitative comparison.}
    \label{fig:VisualComparison}
\end{figure}
Scene 2 was rendered by both engines from the same viewpoint with two lights and shadows enabled to perform qualitative comparisons (see Fig. \ref{fig:VisualComparison}). Improvements made to the reflection model in FIRE-3DV enable an increased visual clarity of the model's details and an overall more physically realistic image. Additionally, the ray-traced shadows in FIRE-3DV are more precise than the AMBF shadows generated with shadow maps. Lastly, the jagged anti-aliasing artifacts at the edges of the geometry in AMBF are virtually non-existent in FIRE-3DV.
\section{Discussion and Future Work}
This paper presents the initial infrastructure required to modernize a dynamic simulation engine, such as AMBF, with a modern graphics API capable of taking advantage of GPU hardware advancements. Most of the features presented in this paper serve as groundwork for future iterations of the engine. For example, vertex buffer addressing, a PBR reflection model, and the acceleration structure build system are all key preliminaries to a full ray-tracing pipeline. Additionally, incorporating a post-processing render pass will easily enable us to implement more advanced image processing effects, such as depth-of-field and bloom. Lastly, integration with the Dear ImGui library is the first step toward a custom scene editor that can enable modifying scenes at runtime.

The planned development of FIRE-3DV is focused on three major goals: (1) the full migration of existing AMBF capabilities to create a more performant, unified framework for robotics simulation, (2) further development of the engine's ray-tracing pipeline to produce physically based and photorealistic simulations, and (3) improvement of the framework's user experience and tooling to minimize the barrier to entry and increase simulation development productivity.

Other potential areas of research and development include GPU-accelerated soft-body dynamics, volumetric smoke simulation from cutting tools, fluid simulation for pooling bodily fluids, physically based endoscope lenses, reduced visual clarity from lens condensation, sub-surface scattering material models for realistic tissue rendering, and image-based lighting from 360\degree~ abdominal high-dynamic-range images. Overall, the proposed Vulkan-based simulation engine lays the foundation to produce more performant and physically accurate simulations in multiple research domains.
\section*{Acknowledgments and Disclosures}
This work was supported in part by NSF AccelNet awards OISE-1927354 and OISE-1927275.
\section*{Supplementary information} 
For more information, visit the project repository at\\
{\footnotesize\url{https://github.com/FIRE-3DV-repositories/FIRE-3DV}}.
\bibliography{main}
\bibliographystyle{IEEEtran}
\end{document}